\documentclass{sigchi}

\usepackage{booktabs}
\usepackage{balance}       
\usepackage{graphics}      
\usepackage{mathptmx}
\usepackage[pdflang={en-US},pdftex]{hyperref}
\usepackage{color}
\usepackage[colorinlistoftodos]{todonotes}
\usepackage{csquotes}
\usepackage{changes}
\usepackage{tabularx}
\usepackage{booktabs}
\usepackage{textcomp}
\usepackage{caption}
\usepackage{subcaption}
\usepackage[none]{hyphenat} 
\usepackage{makecell}
\usepackage{colortbl}
\usepackage{xcolor}

\colorlet{tablerowcolor}{gray!10} 
\newcommand{\rowcol}{\rowcolor{tablerowcolor}}

\usepackage{microtype}        
\usepackage{ccicons}          

\usepackage{todonotes}
\usepackage{makecell}
\setcellgapes{4pt}

\def\plaintitle{Automated Rationale Generation: A Technique for Explainable AI and its Effects on Human Perceptions}

\def\emptyauthor{}
\def\plainkeywords{Authors' choice; of terms; separated; by
  semicolons; include commas, within terms only; required.}

\makeatletter
\def\url@leostyle{%
  \@ifundefined{selectfont}{
    \def\UrlFont{\sf}
  }{
    \def\UrlFont{\small\bf\ttfamily}
  }}
\makeatother
\def\pprw{8.5in}
\def\pprh{11in}

\definecolor{linkColor}{RGB}{6,125,233}
\hypersetup{%
  pdftitle={\plaintitle},
  pdfauthor={\emptyauthor},
  pdfkeywords={\plainkeywords},
  pdfdisplaydoctitle=true, 
  bookmarksnumbered,
  pdfstartview={FitH},
  colorlinks,
  citecolor=black,
  filecolor=black,
  linkcolor=black,
  urlcolor=linkColor,
  breaklinks=true,
  hypertexnames=false
}

\begin{document}

\title{\plaintitle}

\numberofauthors{1}
\author{%
\alignauthor{Upol Ehsan$^{1,2}$, Pradyumna Tambwekar$^{1}$, Larry Chan$^{1}$, Brent Harrison$^{3}$, and Mark Riedl$^{1}$\\
    \affaddr{$^{1}$School of Interactive Computing, Georgia Institute of Technology}\\
    \affaddr{$^{2}$Information Science, Cornell University}\\
    \affaddr{$^{3}$Department of Computer Science, University of Kentucky}\\
    \email{\{ehsanu; ptambwekar3; larrychan; riedl\}@gatech.edu, harrison@cs.kyu.edu}}}

\maketitle

\begin{abstract}
{\em Automated rationale generation} is an approach for real-time explanation generation whereby a computational model learns to translate an autonomous agent's internal state and action data representations into natural language. 
Training on human explanation data can enable agents to learn to generate human-like explanations for their behavior.
%
In this paper, using the context of an agent that plays {\em Frogger}, we describe 
(a) how to collect a corpus of explanations, 
(b) how to train a neural rationale generator to produce 
different styles of rationales, and
(c) how people perceive these rationales.
%
We conducted two user studies. 
The first study establishes the plausibility of each type of generated rationale and situates their user perceptions along the dimensions of \textit{confidence}, \textit{humanlike-ness}, \textit{adequate justification}, and \textit{understandability}.
The second study 
further explores user preferences between the 
generated rationales
with regard to \textit{confidence} in the autonomous agent, communicating \textit{failure} and \textit{unexpected behavior}. 
Overall, we find alignment between the intended differences in features of the generated rationales and the perceived differences by users.
Moreover, context permitting, participants preferred detailed rationales to  
form a stable mental model of the agent's behavior.
%
%
\end{abstract}

\section{Introduction}

{\em Explainable AI} refers to artificial intelligence and machine learning techniques that can provide human understandable justification for their behavior. 
 Explainability is important in situations where human operators work alongside autonomous and semi-autonomous systems because it can help build rapport, confidence, and understanding between the agent and its operator.
In the event that an autonomous system fails to complete a task or completes it in an unexpected way, explanations help the human collaborator 
understand the circumstances that led to the behavior, which also allows the operator to make an informed decision on how to address the behavior. 

Prior work on explainable AI (XAI) has primarily focused on non-sequential problems such as image classification and captioning ~\cite{wang2017residual,xu2015show,you2016image}. 
Since these environments are episodic in nature, the model's output depends only on its input.
In sequential environments, decisions that the agent has made in the past influence future decisions. 
To simplify this, agents often make locally optimal decisions by selecting actions that maximize some discrete notion of expected future reward or utility.
To generate plausible explanations in these environments, the model must unpack this local reward or utility to reason about how current actions affect future actions. On top of that, it needs to communicate the reasoning in a human understandable way, which is a difficult task. 
To address this challenge of human understandable explanation in sequential environments, we introduce the alternative task of rationale generation in sequential environments. 


{\em Automated rationale generation} is a process of producing a natural language explanation for agent behavior {\em as if a human had performed the behavior}~\cite{ehsan2017rationalization}.
The intuition behind rationale generation is that humans can engage in effective communication 
by verbalizing plausible motivations for their action. The communication can be effective even when the verbalized reasoning does not have a consciously accessible neural correlate of 
the decision-making process~\cite{block2007consciousness,block2005two,fodor1994elm}. 
Whereas an explanation can be in any communication modality, rationales are natural language explanations that don't literally expose the inner workings of an intelligent system.
Explanations can be made by exposing the inner representations and data of a system, though this type of explanation may not be accessible or understandable to non-experts.
In contrast, contextually appropriate natural language rationales are 
accessible and intuitive to non-experts, facilitating understanding and communicative effectiveness. 
Human-like communication can also afford human factors advantages such as higher degrees of satisfaction, confidence, rapport, and willingness to use autonomous systems.
Finally, rationale generation is fast, sacrificing 
an accurate view of agent decision-making for real-time response, making it appropriate for real-time human-agent collaboration. 
Should deeper, more grounded and technical
explanations be necessary, rationale generation may need to be supplemented by other explanation or visualization techniques.

In preliminary work~\cite{ehsan2017rationalization} we showed that recurrent neural networks can be used to translate internal state and action representations into natural language. 
That study, however, relied on synthetic natural language data for training. 
In this work, we explore if human-like plausible rationales can be generated using a non-synthetic, natural language corpus of human-produced explanations. 
To create this corpus, we developed a methodology for conducting remote think-aloud protocols \cite{fonteyn1993description}. 
Using this corpus, we then use 
a neural network based on~\cite{ehsan2017rationalization}
to translate an agent's state and action information into natural language rationales, and show how variations in model inputs can produce two different types of rationales. 
Two user studies help us understand the perceived quality of the generated rationales along dimensions of human factors. 
The first study indicates that our rationale generation technique produces plausible and high-quality rationales and explains the differences in user perceptions. 
In addition to understanding user preferences, the second study demonstrates how the intended design behind the rationale types aligns with their user perceptions.

The philosophical and linguistic discourse around the notion of explanations~\cite{miller2017explanation, lipton2001good} is beyond the scope of this paper. 
To avoid confusion, we use the word "rationale" to refer to natural language-based post-hoc explanations that are meant to sound like what a human would say in the same situation. 
We opt for "rationale generation" instead of "rationalization" to signal that the agency lies with the receiver and interpreter (human being) instead of the producer (agent). 
Moreover, the word rationalization may carry a connotation of making excuses~\cite{maruna2006fundamental} for an (often controversial) action, which is another reason why we opt for \textit{rationale generation} as a term of choice.

In this paper, we make the following contributions in this paper:
\begin{itemize}
    \item We present a methodology for collecting high-quality human explanation data based on remote think-aloud protocols. 
    \item We show how this data can be used to configure neural translation models to produce two types of human-like rationales: $\left(1\right)$ concise, localized and $\left(2\right)$ detailed, holistic rationales. We demonstrate the alignment between the intended design of rationale types and the actual perceived differences between them.
    \item We quantify the perceived quality of the rationales and preferences between them, 
    and we use qualitative data to explain these perceptions and preferences.
\end{itemize}

\section{Related Work}
Much of the previous work on explainable AI has focused on {\em interpretability}.
While there is no one definition of interpretability with respect to machine learning models, we view interpretability as a property of machine learned models that dictate the degree to which a human user---AI expert or user---can come to conclusions about the performance of the model on specific inputs.
Some types of models are inherently interpretable, meaning they require relatively little effort to understand. 
Other types of models require more effort to make sense of their performance on specific inputs. 
Some non-inherently interpretable models can be made interpretable in a post-hoc fashion through explanation or visualization.
Model-agnostic post-hoc methods can help to make models intelligible without custom explanation or visualization technologies and without changing the underlying model to make them more interpretable~\cite{ribeiro2016should,yosinski2015understanding}. 

Explanation generation can be described as a form of {\em post-hoc interpretability}~\cite{2016arXiv160603490L, miller2017explanation}; explanations are generated on-demand based on the current state of a model and---potentially---meta-knowledge about how the algorithm works.
An important distinction between interpretability and explanation is that explanation does not elucidate precisely how a model works but aims to give useful information for practitioners and end users.
Abdul et al.~\cite{abdul2018trends} conduct a comprehensive survey on trends in explainable and intelligible systems research.

Our work on rationale generation is a model-agnostic explanation system that works by translating the internal state and action representations of an arbitrary reinforcement learning system into natural language.
Andreas, Dragan, and Klein~\cite{andreas2017translating} describe a technique that translates message-passing policies between two agents into natural language.
An alternative approach to translating internal system representations into natural language is to add explanations to a supervised training set such that a model learns to output a classification as well as an explanation~\cite{codella2018teaching}.
This technique has been applied to generating explanations about procedurally generated game level designs~\cite{guzdial2018explainable}.

Beyond the technology, user perception and acceptance matter because they influence trust in the system, which is crucial to adoption of the technology. 
Established fields such as information systems enjoy a robust array of technology acceptance models such as the Technology Acceptance Model (TAM) \cite{davis1989perceived} and Unified Theory of Acceptance and Use of Technology Model (UTAUT) \cite{venkatesh2003user} whose main goal is to explain variables that influence user perceptions. 
Utilizing dimensions such as perceived usefulness and perceived ease of use, the TAM model aimed to explain prospective expectations about the technological artifacts. 
UTAUT uses constructs like performance expectancy, effort expectancy, etc. to understand technology acceptance. The constructs and measures in these models build on each other. 

In contrast, due to a rapidly evolving domain, a robust and well-accepted user perception model of XAI agents is yet to be developed. 
Until then, we can take inspiration from general acceptance models (such as TAM and UTAUT) and adapt their constructs to understand the perceptions of XAI agents. 
For instance, the human-robot interaction community has used them as basis to understand users' perceptions towards robots~\cite{ezer2009attitudinal, beer2011understanding}. 
While these acceptance models are informative, they often lack sociability factors such as "humanlike-ness".
Moreover, TAM-like models does not account for autonomy in systems, let alone autonomous XAI systems. 
Building on some constructs from TAM-like models and original formative work, we attempt to address the gaps in understanding user perceptions of rationale-generating XAI agents.  

The dearth of established methods combined with the variable conceptions of explanations make evaluation of XAI systems challenging. 
Binns et al.~\cite{binns2018s} use scenario-based survey design~\cite{carroll2000making} and presented different types of hypothetical explanations for the same decision to measure perceived levels of justice. 
One non-neural based network evaluates the usefulness and naturalness of generated explanations~\cite{broekens2010you}.
Rader et al.~\cite{rader2018explanations} use explanations manually generated from content analysis of Facebook's News Feed to study perceptions of algorithmic transparency. 
One key differentiating factor of our approach is that our evaluation is based rationales that are actual system outputs (compared to hypothetical ones). 
Moreover, user perceptions of our system's rationales directly influence the design of our rationale generation technique. 

\begin{figure}[t]
\centering
    \includegraphics[width=1.0\linewidth]{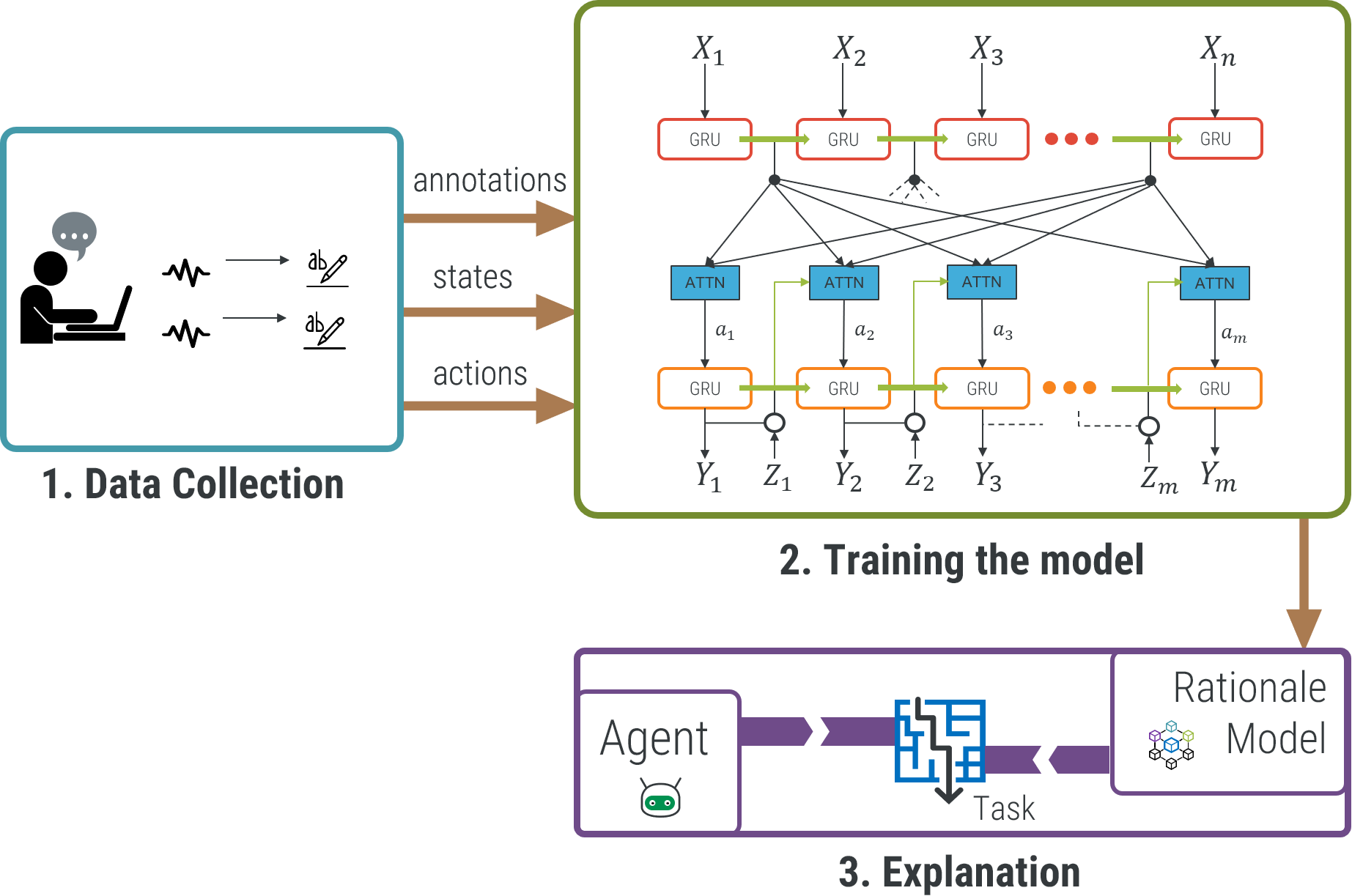}
    \vspace{-1.5\baselineskip}
    \caption{End-to-end pipeline for training a system that can generate explanations.}
\label{fig:end-to-end}
\end{figure}

\section{Learning to Generate Rationales}

We define a \textit{rationale} as an explanation that justifies an action based on how a human would think. 
These rationales do not necessarily reveal the true decision making process of an agent, but still provide insights about why an agent made a decision in a form that is easy for non-experts to understand.

Rationale generation requires translating events in the game environment into natural language outputs. Our approach to rationale generation involves two steps: (1)~collect a corpus of think-aloud data from players who explained their actions in a game environment; and (2)~use this corpus to train an encoder-decoder network to generate plausible rationales for any action taken by an agent (see Figure~\ref{fig:end-to-end}).

We experiment with rationale generation using autonomous agents that play the arcade game, {\em Frogger}.
Frogger is a good candidate for our experimental design of a rationale generation pipeline for general sequential decision making tasks because it is a simple Markovian environment, making it an ideal stepping stone towards a real world environment. 
Our rationale generation technique is agnostic to the type of agent or how it is trained, as long as the representations of states and actions used by the agent can be exposed to the rationale generator and serialized.

\subsection{Data Collection Interface}

There is no readily available dataset for the task of learning to generate explanations. Thus, we developed a methodology to collect live ``think-aloud'' data from players as they played through a game. This section covers the two objectives of our data collection endeavor:
\begin{enumerate}
\item Create a think-aloud protocol in which players provide natural rationales for their actions. 
\item Design an intuitive player experience that facilitates accurate matching of the participants' utterances to the appropriate state in the environment. 
\end{enumerate}

\begin{figure}[t]
\begin{center}
  \includegraphics[width=0.75\linewidth]{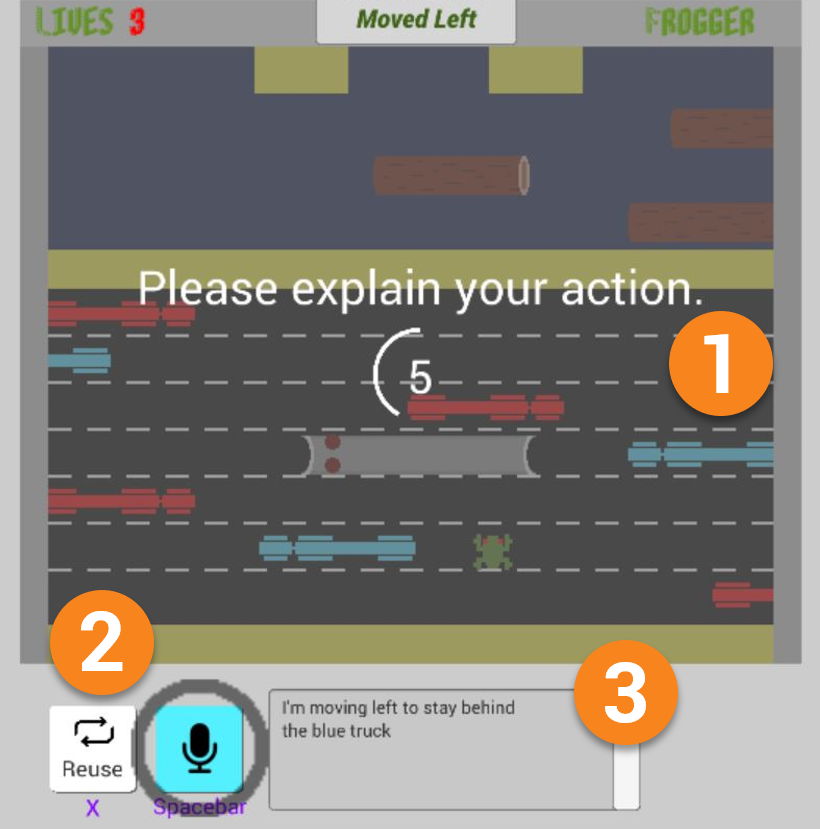}
  \end{center}
  \caption{Players take an action and verbalize their rationale for that action. (1)~After taking each action, the game pauses for 10 seconds. (2)~Speech-to-text transcribes the participant's rationale for the action. (3)~Participants can view their transcribed rationales near real-time and edit if needed.}
  \label{fig:Play}
\end{figure}

To train a rationale-generating explainable agent, we need data linking game states and actions to their corresponding natural language explanations. To achieve this goal, we built a modified version of Frogger in which players simultaneously play the game and also explain each of their actions. 
The entire process is divided into three phases: (1)~A guided tutorial, (2)~rationale collection, and (3)~transcribed explanation review. 

During the guided tutorial~(1), our interface provides instruction on how to play through the game, how to provide natural language explanations, and how to review/modify any explanations they have given. 
This helps ensure that users are familiar with the interface and its use before they begin providing explanations. 

For rationale collection~(2), participants play through the game while explaining their actions out loud in a turn-taking mechanism.
Figure~\ref{fig:Play} shows the game embedded into the explanation collection interface. 
To help couple explanations with actions (attach annotations to concrete game states), the game pauses for 10 seconds after each action is taken.
During this time, the player's microphone automatically turns on and the player is asked to explain their most recent action while a speech-to-text library \cite{github_2017} automatically transcribes the explanation real-time. 
The automatic transcription substantially reduces participant burden as it is more efficient 
than typing an explanation.  
Player can use more or less than the default 10-second pause to collect the explanation. 
Once done explaining, they can view their transcribed text and edit it if necessary.
During pretesting with 14 players, we observed that players often repeat a move for which the explanation is the same as before. 
To reduce burden of repetition, we added a "redo" button that can be used to recycle rationales for consecutive repeated actions.

When the game play is over, players move to transcribed explanation review portion ~(3). Here, they can can step through all the actions-explanation pairs. This stage allows reviewing in both a situated and global context. 

The interface is designed so that no manual hand-authoring/editing of our explanation data was required before 
using it to train our machine learning model. Throughout the game, players have the opportunity to organically edit their own data without impeding their work-flow. This added layer of organic 
editing is crucial in ensuring that we can directly input the collected data into the network with zero manual cleaning. While we use Frogger as a test environment in our experiments, a similar user experience can be designed using other turn-based environments with minimal effort. 

\begin{figure}[t]
  \includegraphics[width=\linewidth]{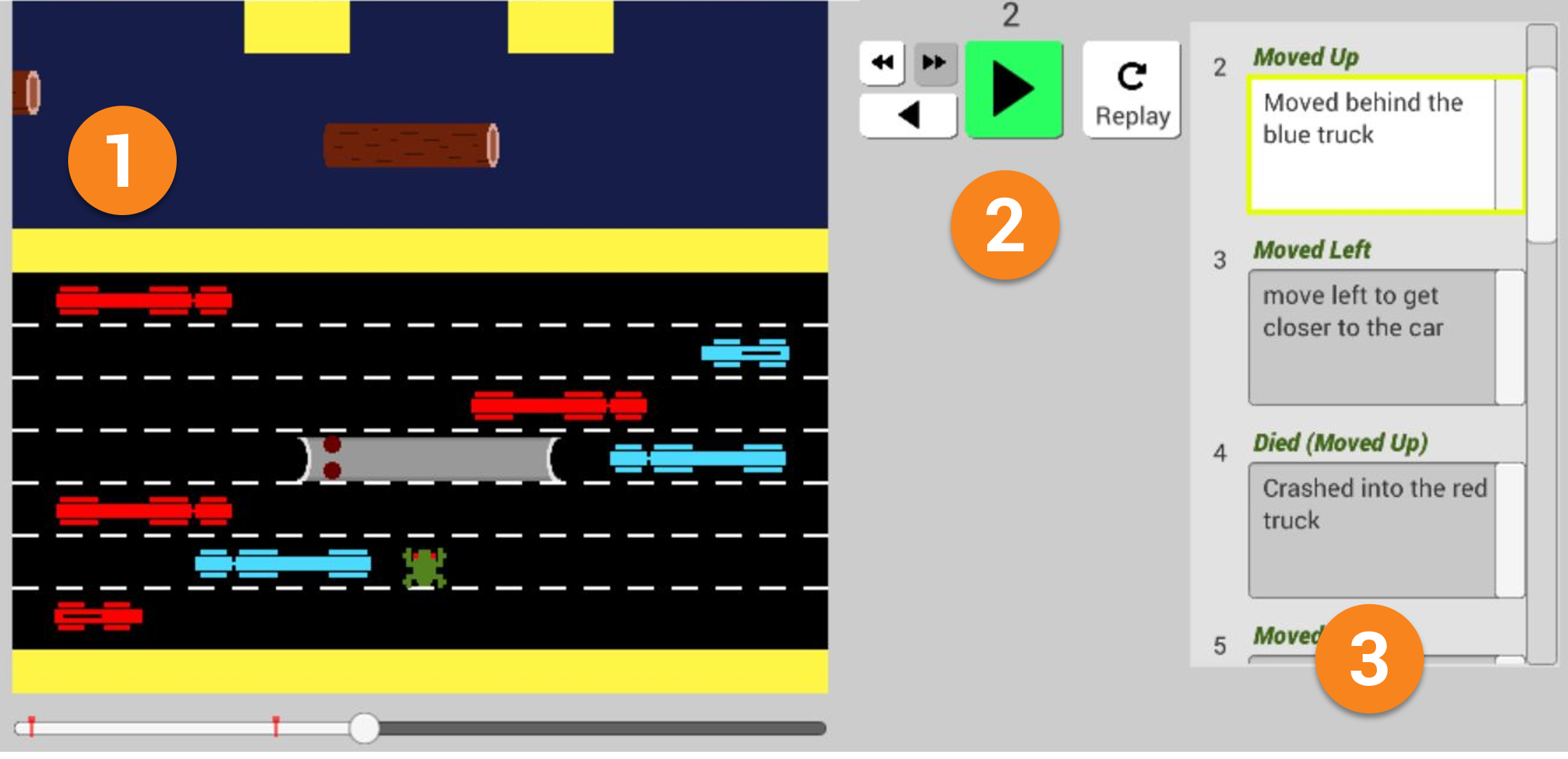}
\caption{Players can step-through each of their action-rationale pairs and edit if necessary. (1)~Players can watch an action-replay while editing rationales. (2)~These buttons control the flow of the step-through process. (3)~The rationale for the current action gets highlighted for review.}
  \label{fig:Review}
\end{figure}

\subsection{Neural Translation Model}

We use an encoder-decoder network~\cite{bahdanau2014neural} to teach our network to generate relevant natural language explanations for any given action. 
These kinds of networks are commonly used for machine translation tasks or dialogue generation, but their ability to understand sequential dependencies between the input and the output make it suitable for our task. 
Our encoder decoder architecture is similar to that used in \cite{ehsan2017rationalization}. 
The network learns how to translate the input game state representation
$X = x_1, x_2, ..., x_n$, comprised of the representation of the game combined with other influencing factors,
into an output rationale as a sequence of words 
$Y = y_1, y_2, ..., y_m$
where $y_i$ is a word.
%
%
Thus our network learns to translate game state and action information into natural language rationales.

The encoder and decoder are both recurrent neural networks (RNN) comprised of Gated Recurrent Unit (GRU) cells since our training process involved a small amount of data.
The decoder network uses an additional attention mechanism~\cite{luong2015effective} to learn to weight the importance of different components of the input with regard to their effect on the output. 

To simplify the learning process, the state of the game environment is serialized into a sequence of symbols where each symbol characterizes a sprite in the grid-based represntation of the world.
To this, we append information concerning Frogger's position, the most recent action taken, and the number of lives the player has left to create the input representation $X$. 
On top of this network structure, we vary the input configurations with the intention of producing varying styles of rationales. 
Empirically, we found that a reinforcement learning agent using tabular $Q$-learning \cite{watkins92} learns to play the game effectively when given a limited window for observation.
Thus a natural configuration for the rationale generator is to give it the same observation window that the agent needs to learn to play.
We refer to this configuration of the rationale generator as {\em focused-view} generator.
This view, however, potentially limits the types of rationales that can be learned since the agent will only be able to see a subset of the full state. Thus we formulated a second configuration that gives the rationale generator the ability to use all information on the board to produce rationales.
We refer to this as {\em complete-view} generator.
An underlying question is thus whether rationale generation should use the same information that the underlying black box reasoner needs to solve a problem or if more information is advantageous at the expense of making rationale generation a harder problem.
In the studies described below, we seek to understand how these configurations affect human perceptions of the agent when presented with generated rationales.

\subsubsection{Focused-view Configuration}
In the \textit{focused-view} configuration, we used a windowed representation of the grid, i.e. only a $7\times7$ window around the Frog was used in the input. 
Both playing an optimal game of Frogger and generating relevant explanations based on the current action taken typically only requires this much local context. 
Therefore providing the agent with only the window around Frogger helps the agent produce explanations grounded in it's neighborhood. 
In this configuration, we designed the inputs such that the network is prone to prioritize short-term planning producing localized rationales instead of long-term planning. 


\subsubsection{Complete-view Configuration}
The \textit{complete-view} configuration is an alternate setup that provides the entire game board as context for the rationale generation.
There are two differences between this configuration and the focused-view configuration.
First, we use the entire game screen as a part of the input. The agent now has the opportunity to learn which other long-term factors in the game may influence it's rationale.
Second, we added noise to each game state to force the network to generalize when learning, reduce the likelihood that spurious correlations are identified, and to give the model equal opportunity to consider factors from all sectors of the game screen.
In this case noise was introduced by replacing input grid values with dummy values. For each grid element, there was a $20\%$ chance that it would get replaced with a dummy value.
Given the input structure and scope, this configuration should prioritize rationales that exhibit long-term planning and consider the broader context.

  

\begin{table}[h]
  \caption{Examples of \textit{focused-view} vs  \textit{complete-view} rationales generated by our system for the same set of actions.}
  \label{tab:components}
  \begin{tabular}{p{0.1\columnwidth}p{0.4\columnwidth}p{0.4\columnwidth}}
    \toprule
    {\bf Action} & {\bf Focused-view} & {\bf Complete-view}\\
    \midrule
    Right & I had cars to the left and in front of me so I needed to move to the right to avoid them. & I moved right to be more centered. This way I have more time to react if a car comes from either side. \\
    \rowcol Up & The path in front of me was clear so it was safe for me to move forward.  & I moved forward making sure that the truck won\textquotesingle t hit me so I can move forward one spot. \\
    Left & I move to the left so I can jump onto the next log. & I moved to the left because it looks like the logs and top or not going to reach me in time, and I\textquotesingle m going to jump off if the law goes to the right of the screen. \\
    \rowcol Down & I had to move back so that I do not fall off. & I jumped off the log because the middle log was not going to come in time. So I need to make sure that the laws are aligned when I jump all three of them. \\
  \bottomrule
\end{tabular}
\end{table}

\section{Perception Study: Candidate vs. Baseline Rationales}
In this section, we 
assess whether the rationales generated using our technique are plausible and explore 
how humans perceive them along various dimensions of human factors.
For our rationales to be plausible we would expect that human users indicate a strong preference for rationales generated by our system (either configuration) over those generated by a baseline rationale generator.
We also compare them to exemplary human-produced explanations to get a sense for how far from the upper bound we are.

This study aims to achieve two main objectives.
First, it seeks to confirm the hypothesis that humans prefer rationales generated by each of the configurations over randomly selected rationales across all dimensions.
While this baseline is low, it establishes that rationales generated by our technique are not nonsensical. 
We can also measure the distance from the upper-bound (exemplary human rationales) for each rationale type. 
Second, we attempt to understand the underlying components that influence 
the perceptions of the generated rationales along four dimensions of human factors: {\em confidence}, {\em human-likeness}, {\em adequate justification}, and {\em understandability}.

\subsection{Method}

To gather the training set of game state annotations, we deployed our data collection pipeline on {\em Turk Prime}~\cite{litman2017turkprime}. 
From 60 participants
we collected over 2000 samples of human actions in Frogger coupled with natural language explanations. The average duration of this task was around 36 minutes. 
%
The parallel corpus of the collected game state images and  natural language explanations was used to train the encoder-decoder network. 
Each RNN in the encoder and the decoder was parameterized with GRU cells with a hidden vector size of 256. 
The entire encoder-decoder network was trained for 100 epochs.

For the perception user study, we collected both within-subject and between-subject data.
We recruited 128 participants, split into two equal experimental groups through {\em TurkPrime}: Group 1 (age range = 23 - 68 years, M = 37.4, SD = 9.92) and Group 2 (age range = 24 - 59 years, M = 35.8, SD= 7.67). 
On average, the task duration was approximately 49 minutes.
46\% of our participants were women, and the 93\% of participants were self-reported as from the United States while the remaining 7\% of participants were self-reported as from India.

\begin{figure}
\centering
  \includegraphics[width=1.0\linewidth]{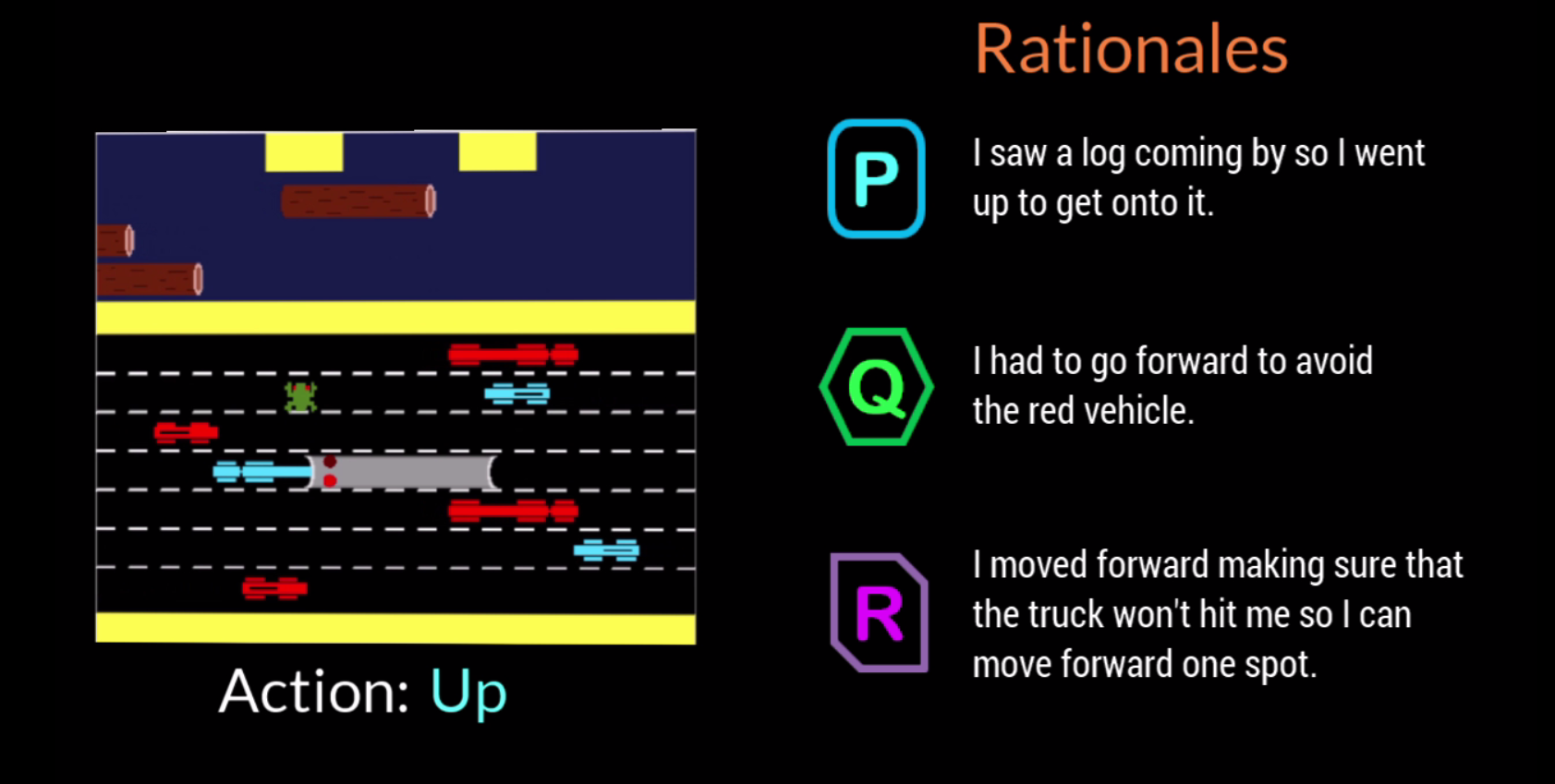}
  \caption{Screenshot from user study (setup 2) depicting the action taken and the rationales: \textit{P = Random, Q = Exemplary, R = Candidate}}
  \label{fig:Study}
\end{figure}

All participants watched a counterbalanced series of five videos.
Each video depicted an action taken by Frogger accompanied by three different types of rationales that justified the action (see Figure~\ref{fig:Study}).
Participants rated each rationale on a labeled, 5-point, bipolar Likert-scale along 4 perception dimensions (described below). 
Thus, each participant provided 12 ratings per action, leading to 60 perception ratings for five actions. 
Actions collected from human players comprised the set of Frogger's actions. These actions were then fed into the system to generate rationales to be evaluated in the the user studies. 
In order to get a balance between participant burden, fatigue, the number of actions, and regions of the game, we pretested with 12 participants. 
Five actions was the limit beyond which participants' fatigue and burden substantially increased. 
Therefore, we settled on five actions (up (twice), down, left, and right) in the major regions of the game-- amongst the cars, at a transition point, and amongst the logs. 
This allowed us to test our rationale generation configurations in all possible action-directions in all the major sections of the game. 

The study had two identical experimental conditions, differing only by type of \textit{candidate rationale}. 
Group 1 evaluated the \textit{focused-view} rationale while Group 2 evaluated the \textit{complete-view} rationales.
In each video, the action was accompanied by three rationales generated by three different techniques (see Figure~\ref{fig:Results_1}):  
\begin{itemize}
\item The \textit{exemplary rationale} is the rationale from our corpus that 3 researchers unanimously agreed on as the best one for a particular action. Researchers independently selected rationales they deemed best and iterated until consensus was reached.
This is provided as an upper-bound for contrast with the next two techniques.
\item  The \textit{candidate rationale} is the rationale produced by our network, either the focused-view or complete-view configuration. 

\item The \textit{random rationale} is a randomly chosen rationale from our corpus.
\end{itemize}


\noindent
For each rationale, participants used a 5-point Likert scale to rate their endorsement of each of following four statements, which correspond to four dimensions of interest. 

\begin{enumerate}
    \item \textit{Confidence:} This rationale makes me confident in the character's ability to perform it's task.
    \item \textit{Human-likeness: } This rationale looks like it was made by a human.
    \item \textit{Adequate justification:} This rationale adequately justifies the action taken.
    \item \textit{Understandability:} This rationale helped me understand why the agent behaved as it did.
\end{enumerate}

Response options on a clearly labeled bipolar Likert scale ranged from "strongly disagree" to "strongly agree". In a mandatory free-text field, they explained their reasoning behind the ratings for a particular set of three rationales. After answering these questions, they provided demographic information.

These four dimensions emerged from an iterative filtering process that included preliminary testing of the study, informal interviews with experts and participants, and a literature review on robot and technology acceptance models. Inspired by the acceptance models, we created a set of dimensions that were contextually appropriate for our purposes. 

Direct one-to-one mapping from existing models was not feasible, given the novelty and context of the Explainable AI technology.
We adapted \textit{confidence}, a dimension that impacts trust in the system \cite{kaniarasu2013robot}, from constructs like performance expectancy \cite{venkatesh2003user} (from UTAUT) and robot performance~\cite{beer2011understanding, chernova2009confidence}.
\textit{Human-likeness}, central to generating human-centered rationales, was inspired from sociability and anthropomorphization factors from HRI work on robot acceptance [\cite{nass1994machines,nass1996can,nass2000machines}. 
Since our rationales are justificatory in nature, \textit{adequate justification} is a reasonable measure of output quality (transformed from TAM).
Our rationales also need to be \textit{understandable}, which can signal perceived ease of use (from TAM).

\subsection{Quantitative Analysis}

We used a multi-level model to analyze our data.
All variables were within-subjects except for one: whether the candidate style was focused-view (Group 1) or complete-view (Group 2).  
This was a between-subject variable.

\begin{figure}[t]
\subcaptionbox{Focus-View condition.\label{fig:Results_1a}}{\includegraphics[width=1.0\linewidth]{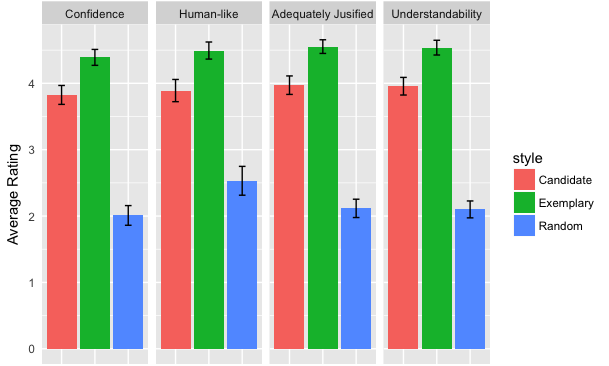}}\hfill
\subcaptionbox{Complete-View condition.\label{fig:Results_1b}}{\includegraphics[width=1.0\linewidth]{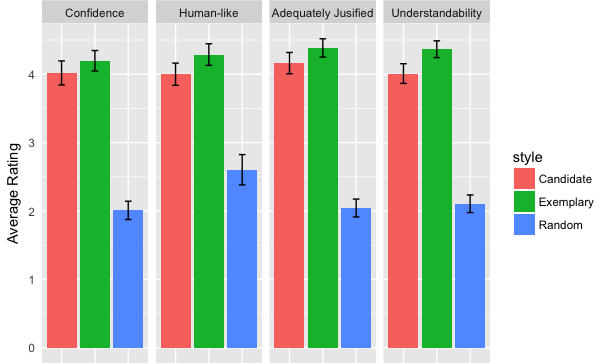}}\hfill
   \caption{Human judgment results. }
  \label{fig:Results_1}
\end{figure}

There were significant main effects of rationale style ($\chi^2\left(2\right) = 594.80, p<.001$) and dimension ($\chi^2\left(2\right) = 66.86, p<.001$) on the ratings. 
The main effect of experimental group was not significant ($\chi^2\left(1\right) = 0.070, p=0.79$).
Figure~\ref{fig:Results_1} shows the average responses to each question for the two different experimental groups.
Our results support our hypothesis that rationales generated with the \textit{focused-view} generator and the \textit{complete-view} generator were judged significantly better across all dimensions than the random baseline
($b=1.90, t\left(252\right)=8.09,p<.001$). 
In addition, exemplary rationales were judged significantly higher than candidate rationales.
Though there were significant differences between each kind of candidate rationale and the exemplary rationales, those differences were not the same. 
The difference between the \textit{focused-view} candidate rationales and exemplary rationales were significantly \textit{greater} than the difference between \textit{complete-view} candidate rationales and exemplary rationales ($p=.005$). 
Surprisingly, this was because the exemplary rationales were rated {\rm lower} in the presence of complete-view candidate rationales ($t\left(1530\right)=-32.12,p<.001$).
Since three rationales were presented simultaneously in each video, it is likely that participants were rating the rationales relative to each other. 
We also observe that the \textit{complete-view} candidate rationales received higher ratings in general than did the \textit{focused-view} candidate rationales ($t\left(1530\right)=8.33,p<.001$).
%

In summary, we have confirmed our hypothesis that both configurations produce rationales that perform significantly better than the \textit{random} baseline across all dimensions. 

\subsection{Qualitative Findings and Discussion}


In this section, we look at the open-ended responses provided by our participants to better understand the criteria that participants used when making judgments about the \textit{confidence, human-likeness, adequate justification,} and \textit{understandability} of generated rationales. 
These situated insights augment our understanding of rationale generating systems, enabling us to design better ones in the future.


We analyzed the open-ended justifications participants provided using a combination of thematic analysis \cite{aronson1994pragmatic} and grounded theory \cite{strauss1994grounded}. 
We developed codes that addressed different types of reasonings behind the ratings of the four dimensions under investigation. 
Next, the research team clustered the codes under emergent themes, which form the underlying \textit{components} of the dimensions. Iterating until consensus was reached, researchers settled on the five most relevant components: (1)~\textit{Contextual Accuracy}, (2)~\textit{Intelligibility}, (3)~\textit{Awareness},
(4)~\textit{Relatability}, and (5)~\textit{Strategic Detail} (see Table \ref{tab:components}).
At varying degrees, multiple components influence more than one dimension; that is, there isn't a mutually exclusive one-to-one relationship between components and dimensions. 

We will now share how these components influence the dimensions of the human factors under investigation.  
When providing examples of our participants' responses, we will use P$1$ to refer to participant 1, P$2$ for participant 2, etc. 
To avoid priming during evaluation, we used letters (e.g., A, B, C, etc.) to refer to the different types of rationales. 
For better comprehension, we have substituted the letters with appropriate rationale--focused-view, complete-view, or random-- while presenting quotes from participants below.  

\subsubsection{Confidence (1)} This dimension gauges the participant's faith in the agent's ability to successfully complete it's task and has \textit{contextual accuracy},  \textit{awareness}, \textit{strategic detail}, and \textit{intelligibility} as relevant components. 
With respect to \textit{contextual accuracy}, rationales that displayed ``\ldots recognition of the environmental conditions and [adaptation] to the conditions'' (P22) were a positive influence, while redundant information such as  ``just stating the obvious'' (P42) hindered confidence ratings. 



\begin{table}[h]
  \caption{Descriptions for the emergent \textit{components} underlying the human-factor \textit{dimensions} of the generated rationales.}
  \label{tab:components}
  \begin{tabular}{p{0.32\columnwidth}p{0.58\columnwidth}}
    \toprule
    {\bf Component} & {\bf Description}\\
    \midrule
    Contextual Accuracy & Accurately describes pertinent events in the context of the environment.\\
    \rowcol Intelligibility & Typically error-free and is coherent in terms of both grammar and sentence structure.\\
    Awareness & Depicts and adequate understanding of the rules of the environment.\\
    \rowcol Relatability & Expresses the justification of the action in a relatable manner and style.\\
    Strategic Detail & Exhibits strategic thinking, foresight, and planning.\\
  \bottomrule
\end{tabular}
\end{table}

Rationales that showed \textit{awareness} of ``upcoming dangers and what the best moves to make \ldots [and] a good way to plan'' (P17) inspired confidence from the participants. 
In terms of \textit{strategic detail}, rationales that showed "\ldots long-term planning and ability to analyze information" (P28) yielded higher confidence ratings compared to those that were "short-sighted and unable to think ahead" (P14) led to lower perceptions of confidence. 

\textit{Intelligibility} alone, without \textit{awareness} or \textit{strategic detail}, was not enough to yield high confidence in rationales. However, rationales that were not \textit{intelligible} (unintelligible) or coherent had a negative impact on participants' confidence:
\begin{displayquote}
The [random and focused-view rationales] include major mischaracterizations of the environment because they refer to an object not present or wrong time sequence, so I had very low confidence. (P66)
\end{displayquote}


\subsubsection{Human-likeness (2)} \textit{Intelligibility, relatability,} and \textit{strategic detail} are components that influenced participants' perception of the extent to which the rationales were made by a human.
Notably, \textit{intelligibility} had mixed influences on the human-likeness of the rationales as it depended on what participants thought ``being human'' entailed. 
Some conceptualized humans as fallible beings and rated rationales with errors more \textit{humanlike} because rationales ``with typos or spelling errors \ldots seem even more likely to have been generated by a human" (P19). 
Conversely, some thought error-free rationales must come from a human, citing that a ``computer just does not have the knowledge to understand what is going on'' (P24).

With respect to \textit{relatability}, rationales were often perceived as more human-like when participants felt that ``it mirrored [their] thoughts'' (P49), and ``\ldots [laid] things out in a way that [they] would have'' (P58). Affective rationales had high \textit{relatability} because they ``express human emotions including hope and doubt'' (P11). 

\textit{Strategic detail} had a mixed impact on human-likeness just like \textit{intelligibility} as it also depended on participants' perception of critical thinking and logical planning. Some participants associated ``\ldots critical thinking [and ability to] predict future situations" (P6) with human-likeness whereas others associated logical planning with non-human-like, but computer-like rigid and algorithmic thinking process flow.

\subsubsection{Adequate Justification (3)} This dimension unpacks the extent to which participants think the rationale adequately justifies the action taken and is influenced by \textit{contextual accuracy}, and \textit{awareness}. 
Participants downgraded rationales containing low levels of \textit{contextual accuracy} in the form of irrelevant details. As P11 puts it: 
\begin{displayquote}
The [random rationale] doesn't pertain to this situation. [The complete-view] does, and is clearly the best justification for the action that Frogger took because it moves him towards his end goal. 
\end{displayquote}

Beyond \textit{contextual accuracy},  rationales that showcase \textit{awareness} of surroundings score high on the \textit{adequate justification} dimension. For instance, P11 rated the \textit{random} rationale low because it showed  ``no awareness of the surroundings''. For the same action, P11 rated \textit{exemplary} and \textit{focused-view} rationales high because each made the participant  ``believe in the character's ability to judge their surroundings.''

\subsubsection{Understandability (4)} 
For this dimension, components such as \textit{contextual accuracy} and \textit{relatability} influence participants' perceptions of how much the rationales helped them understand the motivation behind the agent's actions. 
In terms of \textit{contextual accuracy}, 
many expressed how the contextual accuracy, not the length of the rationale, mattered when it came to understandability.  
While comparing understandability of the \textit{exemplary} and \textit{focused-view} rationales, P41 made a notable observation:
\begin{displayquote}
The [exemplary and focused-view rationale] both described the activities/objects in the immediate vicinity of the frog.  However, [exemplary rationale (typically lengthier than focused)] was not as applicable because the [focused-view] rationale does a better job of providing contextual understanding of the action.
\end{displayquote}

Participants put themselves in the agent's shoes and evaluated the understandability of the rationales based on how \textit{relatable} they were. 
In essence, some asked ``Are these the same reasons I would [give] for this action?'' (P43). 
The more relatable the rationale was, the higher it scored for understandability. 

In summary, the first study establishes the plausibility of the generated rationales (compared to baselines) and their user perceptions. 
However, this study does not provide direct comparison between the two configurations. 

\section{Preference Study: Focused-- vs. Complete--View Rationales}
The preference study puts the rationales in direct comparison with each other. 
This study 
It achieves two main purposes. First, it aims to validate the alignment between the intended design of rationale types and the actual perceived differences between them. 
We collect qualitative data on how participants perceived rationales produced by our \textit{focused-view} and \textit{complete-view} rationale generator.
Our expert observation is that the \textit{focused-view} configuration results in concise and localized rationales whereas the \textit{complete-view} configuration results in detailed, holistic rationales. 
We seek to determine whether na\"ive users who are unaware of which configuration produced a rationale also describe the rationales in this way. 
Second, we seek to understand how and why the preferences between the two styles differed along three dimensions: {\em confidence}, {\em failure}, and {\em unexpected behavior}.


\subsection{Method}
Using similar methods to the first study, we recruited and analyzed the data from 65 people (age range = 23 - 59 years, M = 38.48, SD = 10.16). 57\% percent of the participants were women with 96\% of the participants self-reporting the United States and 4\% self-reporting India as countries they were from. Participants from our first study could not partake in the second one. The average task duration was approximately 46 minutes.



The only difference in the experimental setup between perception and the preference study is the comparison groups of the rationales. 
In this study, participants judged the same set of
\textit{focused-} and \textit{complete-view} rationales, however instead of judging each style against two baselines, participants evaluate the \textit{focused-} and \textit{complete-view} rationales in direction comparison with each other. 

Having watched the videos and accompanying rationales, participants responded to the following questions comparing both configurations: 
\begin{enumerate}
    \item \textbf{Most important difference}: What do you see as the most important difference? Why is this difference important to you?
    \item \textbf{Confidence}: Which style of rationale makes you more confident in the agent's ability to do its task? Was it system A or system B? Why?
    \item \textbf{Failure}: If you had a companion robot that had just made a mistake, would you prefer that it provide rationales like System A or System B? Why? 
    \item \textbf{Unexpected Behaviour}: If you had a companion robot that took an action that was not wrong, but unexpected from your perspective, would you prefer that it provides rationales like System A or System B? Why? 
\end{enumerate}


We used a similar to the process of selecting dimensions in this study as we did in the first one. 
\textit{Confidence} is crucial to trust especially when failure and unexpected behavior happens \cite{chernova2009confidence, kaniarasu2013robot}. 
Collaboration, tolerance, and perceived intelligence are affected by the way autonomous agents and robots communicate \textit{failure} and \textit{unexpected behavior} \cite{desai2013impact,kwon2018expressing,lee2010gracefully,mirnig2017err}. 


\begin{table}[h]
  \caption{Tally of how many preferred the \textit{focused-view} vs. the \textit{complete-view} for the three dimensions.}
  \label{tab:components}
  \begin{tabular}{ccc}
    \toprule
    {\bf Question} & {\bf Focused-view} & {\bf Complete-view}\\
    \midrule
    Confidence & 15 & 48\\
    \rowcol Failure & 17 & 46\\
    Unexpected Behaviour & 18 & 45\\
  \bottomrule
\end{tabular}
\end{table}
\subsection{Quantitative Analysis}
In order to determine whether the preferences significantly favored one style or the other, we conducted  
the Wilcoxon signed-rank test. It showed that preference for the \textit{complete-view} rationale was significant in all three dimensions.
Confidence in the \textit{complete-view} rationale was significantly greater than in the \textit{focused-view}  ($p<.001$). 
Similarly, preference for a \textit{complete-view} rationales from an agent that made a mistake was significantly greater than for \textit{focused-view} rationales ($p<.001$).
Preference for \textit{complete-view} rationales from an agent that made a mistake was also significantly greater than for \textit{focused-view} rationales ($p<.001$).

\subsection{Qualitative Findings and Discussion}
In this section, similar to the first study, we share insights gained from the open-ended responses to reveal the underlying reasons behind perceptions of the \textit{most important difference} between the two styles. We also unpack the reasoning behind the quantitative ranking preferences for \textit{confidence} in the agent's ability to do its task and communication preferences for \textit{failure} and \textit{unexpected behavior}. In this analysis, the interacting \textit{components} that influenced the dimensions of human factors in the first study return (see Table \ref{tab:components}). In particular, we use them as analytic lenses to highlight the trade-offs people make when expressing their preferences and the reasons for the perceived differences between the styles. 

These insights bolster our situated understanding of the differences between the two rationale generation techniques and assist to verify if the intended design of the two configurations aligns with the perceptions of them. In essence, did the design succeed in doing what we set out to do? We analyzed the open-ended responses in the same manner as the first study. We use the same nomenclature to refer to participants. 

\subsubsection{Most Important Difference (1)}
Every participant 
indicted that
the \textit{level of detail and clarity} (P55)
differentiated the rationales. 
Connected to the level of detail and clarity is the perceived \textit{long-} vs. \textit{short-term} planning exhibited by each rationale. 
Overall, participants felt that the \textit{complete-view} rationale showed better levels of \textit{strategic detail}, \textit{awareness}, and \textit{relatability} with human-like justifications, whereas the \textit{focused-view} exhibited better \textit{intelligibility} with easy-to-understand rationales. 
The following quote 
illustrates
the trade-off between succinctness, 
which hampers comprehension of higher-order goals, and broadness, which can be perceived as less 
focused:

\begin{displayquote}
The [focused-view rationale] is extraordinarily vague and focused on the raw mechanics of the very next move \ldots [The complete-view] is more broad and less focused, but takes into account \textit{the entire picture}. So I would say the most important difference is the \textit{scope of events} that they take into account while making justifications [emphasis added] (P24)
\end{displayquote}

Beyond trade-offs, this quote highlights a powerful validating point: without any knowledge beyond what is shown on the video, the participant pointed out how the \textit{complete-view} rationale appeared to consider the "entire picture" and how the "scope of events" taken into account was the main difference. 
The participant's intuition precisely aligns with the underlying network configuration design and our 
research 
intuitions.
Recall that the \textit{complete-view} rationale
was generated using the entire environment or "picture" whereas the \textit{focused-view} was generated using a windowed input.  

In prior sections, we 
speculated on the effects of the network configurations. We expected the \textit{focused-view} version 
to produce 
succinct, localized rationales that concentrated on the short-term. We expected the \textit{complete-view} version 
to produce detailed, broader rationales that focused on the larger picture and long-term planning. 
The findings of this experiment are the first validation that the outputs reflect the intended designs.  
The strength of this validation was enhanced by the many descriptions of our intended attributes, given in free-form by participants who were naive to our network designs.

Connected to the level of detail and clarity is the perception of \textit{short-} vs \textit{long-term} thinking from the respective rationales. 
In general, participants regarded the \textit{focused-view} rationale having low levels of \textit{awareness} and \textit{strategic detail}. 
They felt 
that this agent "\ldots focus[ed] only on the current step" (P44), which was perceived depicting as thinking "\ldots in the spur of the moment" (P27), giving the perception of short-term and simplistic thinking.  
On the other hand, the \textit{complete-view} rationale appeared to "\ldots try to think it through" (P27), exhibiting long-term thinking as it appears to "\ldots think forward to broader strategic concerns."(P65)  One participant sums it up nicely: 
\begin{displayquote}
The [focused-view rationale] focused on the immediate action required. [The complete-view rationale] took into account the current situation, [but] also factored in what the next move will be and what dangers that move poses. The [focused-view] was more of a short term decision and [complete-view] focused on both short term and long term goals and objectives. (P47)
\end{displayquote}

We will notice how these differences in perception impact other dimensions such as confidence and communication preferences for failure and unexpected behavior. 
\subsubsection{Confidence (2)}
Participants had more confidence in the agent's ability to do its task if the rationales exhibited high levels of \textit{strategic detail} in the form of long-term planning, \textit{awareness} via expressing knowledge of the environment, and \textit{relatability} through humanlike expressions. They associated \textit{conciseness} with confidence when the rationales did not need to be detailed given the context of the (trivial) action. 

The \textit{complete-view} rationale inspired more confidence because participants perceived agents with long-term planning and high
\textit{strategic detail} as being "more predictive" and intelligent than their counterparts. Participants felt more at ease because "\ldots knowing what [the agent] was planning to do ahead of time would allow me to catch mistakes earlier before it makes them." (P31) As one participant put it:

\begin{displayquote}
The [complete-view rationale] gives me more confidence \ldots because it thinks about future steps and not just the steps you need to take in the moment. [The agent with focused-view] thinks more simply and is prone to mistakes. (P13)
\end{displayquote}

Participants felt that rationales that exhibited a better understanding of the environment, and thereby better \textit{awareness}, resulted in higher confidence scores. Unlike the \textit{focused-view} rationale that came across as "a simple reactionary move \ldots [the \textit{complete-view}] version demonstrated a more thorough understanding of the entire field of play." (P51) In addition, the \textit{complete-view} was more \textit{relatable} and confidence-inspiring "because it more closely resemble[d] human judgment" (P29). 

\subsubsection{Failure (3)}
When an agent or a robot fails, the information from the failure report is mainly used to fix the issue. To build a mental model of the agent, participants preferred \textit{detailed} rationales with solid \textit{explanatory power} stemming from \textit{awareness} and \textit{relatability}. The mental model could facilitate proactive and preventative care. 

The \textit{complete-view} rationale, due to relatively high \textit{strategic detail}, was preferable in communicating failure because participants could "\ldots understand the full reasoning behind the movements."(P16) Interestingly, \textit{detail} trumped \textit{intelligibility} in most circumstances. Even if the rationales had some grammatical errors or were a "\ldots little less easy to read, the details made up for it." (P62) 

However, detailed rationales are not always a virtue. Simple rationales have the benefit of being easily understandable to humans, even if they cause humans to view the agent as having limited understanding capabilities. Some participants appreciated \textit{focused-view} rationales because they felt "it would be easier to figure out what went wrong by focusing on one step at a time."  


Explanatory power, specifically how events are communicated, is related to \textit{awareness} and \textit{relatability}. Participants preferred relatable agents that "\ldots would talk to [them] like a person would."(P11) They expressed the need to develop a mental model, especially to "\ldots see how [a robot's] mind might be working"(P1), to effectively fix the issue. The following participant neatly summarizes the dynamics:
\begin{displayquote}
I'd want [the robot with complete-view] because I'd have a better sense of the steps taken that lead to the mistake. I could then fix a problem within that reasoning to hopefully avoid future mistakes. The [focused-view rationale] was just too basic and didn't give enough detail. (P8)
\end{displayquote}
\subsubsection{Unexpected Behavior (4)}
Unexpected behavior that is not failure makes people want to know the "why?" behind the action, especially to understand the expectancy violation. 
As a result, participants preferred rationales with transparency so that they can understand and trust the robot in a situation where expectations are violated. 
In general, preference was for adequate levels of \textit{detail} and \textit{explanatory power} that could provide "\ldots more diagnostic information and insight into the robot's thinking processes."(P19) 
Participants wanted to develop mental models of the robots so they could understand the world from the robot's perspective. 
This diagnostic motivation for a mental model is different from the re-programming or fixing needs in cases of failure. 

The \textit{complete-view} rationale, due to adequate levels of \textit{strategic detail}, made participants more confident in their ability to follow the thought process and get a better understanding of the expectancy violation. One participant shared:
\begin{displayquote}
The greater clarity of thought in the [complete-view] rationale provides a more thorough picture \ldots, so that the cause of the unexpected action could be identified and explained more easily. (P51)
\end{displayquote}
With this said, where possible without sacrificing transparency, participants welcomed simple rationales that "anyone could understand, no matter what their level of education was."(P2)
This is noteworthy because the expertness level of the audience is a key concern when making accessible AI-powered technology where designers need to strike a balance between detail and succinctness.

Rationales exhibiting strong explanatory power, through \textit{awareness} and \textit{relatability}, helps to situate the unexpected behavior in an understandable manner. 
Participants preferred the \textit{complete-view} rationale's style of communication because of increased transparency:
\begin{displayquote}
I prefer [the complete-view rationale style] because \ldots I am able to get a much better picture of why it is making those decisions. (P24)
\end{displayquote}

Despite similarities in the communication preferences for failure and unexpected behavior, there are differences in underlying reasons. As our analysis suggests, the mental models are desired in both cases, but for different reasons. 
\section{Design Lessons and Implications} 
The situated understanding of the \textit{components} and \textit{dimensions} give us a powerful set of actionable insights that can help us design better human-centered, rationale-generating, autonomous agents.
As our analysis reveals, context is king. 
Depending on the context, we can tweak the input type to generate \textit{rationale sytles} that meet the needs of the task or agent persona; for instance, a companion agent that requires high \textit{relatability} for user engagement. 
We should be mindful when optimizing for a certain dimension as each component comes with costs. 
For instance, conciseness can improve \textit{intelligibility} and overall \textit{understandability} but comes at the cost of \textit{strategic detail}, which can hurt \textit{confidence} in the agent.
We can also engineer systems such that multiple network configurations act as modules. 
For instance, if we design a companion agent or robot that interacts with a person longitudinally, the \textit{focused-view} configuration can take over 
when short and simple rationales are required. 
The \textit{complete-view} configuration or a hybrid one can be activated when communicating failure or unexpected behavior.

As our preference study shows, we should not only be cognizant about the level of detail, but also why the detail is necessary, especially while communicating failure and unexpected behavior.
For instance, 
failure-reporting,
in a mission critical task (such as search and rescue), would have different requirements for \textit{strategic detail} and \textit{awareness}, compared to 
"failure" reporting in a 
less-defined, more creative task like making music. 
While the focus of this paper is on textual rationale generation, rationales can be complementary to other types of explanations; for instance, a multi-modal system can combine visual cues with textual rationales to provide better contextual explanations for an agent's actions.  


\section{Limitations and Future Work}
While these results are promising, there are several limitations in our approach that need to be addressed in future work. 
First, our current system, by intention and design, lacks interactivity; users cannot contest a rationale or ask the agent to explain in a different way. 
To a get a formative understanding, we kept the design as straight-forward as possible.
Now that we have a baseline understanding, we can vary along the dimension of interactivity for the next iteration. 
For instance, contestability, the ability to either reject a reasoning or ask for another one, which has shown to improve user satisfactions \cite{hirsch2017designing,dietvorst2016overcoming} can be incorporated in the future.
Second, our data collection pipeline is currently designed to work with discrete-action games that have natural break points where the player can be asked for explanations. 
In continuous-time and -action environments, we must determine how to collect the necessary data without being too intrusive to participants. 
Third, all conclusions about our approach were formed based on one-time interactions with the system. 
To better control for potential novelty effects that rationales could have, we need to deploy our system in a longitudinal task setting. 
Fourth, to understand the feasibility of our system in larger state-action spaces, we would need to study the scalability by addressing the question of how much data is needed based on the size of environment. 
Fifth, not all mistakes are created equal. Currently, the perception ratings are averaged where everything is equally weighted. For instance, a mistake during a mission critical step can lead to higher fall in confidence than the same mistake during a non-critical step. To understand the relative costs of mistakes, we need to further investigate the relationship between context of the task and the cost of the mistake.  

\section{Conclusions}
While explainability has been successfully introduced for classification and captioning tasks, sequential environments offer a unique challenge for generating human understandable explanations.
The challenge stems from multiple complex factors, such as temporally connected decision-making, that contribute to making decisions in these environments.
In this paper, we introduce \textit{automated rationale generation} as a concept and explore how justificatory explanations from humans can be used to train systems to produce human-like explanations in sequential environments. 
To facilitate this work, we also introduce a pipeline for automatically gathering a parallel corpus of states annotated with human explanations. 
This tool enables us to systematically gather high quality data for training purposes.
We then use this data to train a model that uses machine translation technology to generate human-like rationales in the arcade game, {\em Frogger}. 

Through a mixed-methods approach in evaluation, we establish the plausibility of the generated rationales and describe how intended design of  rationale  types  lines up with the  actual  user perceptions of them. 
We also get contextual understanding of the underlying dimensions and components that influence human perception and preferences of the generated rationales. 
By enabling autonomous agents to communicate about the motivations for their actions, we envision a future where explainability not only improves human-AI collaboration, but does so in a human--centered and understandable  manner.

\section{Acknowledgements}
This work was partially funded under ONR grant number N00014141000. We would like to thank Chenghann Gan and Jiahong Sun for their valuable contributions to the development of the data collection pipeline. We are also grateful to the feedback from anonymous reviewers that helped us improve the quality of the work.
\balance{}

\balance{}

\bibliographystyle{SIGCHI-Reference-Format}
\bibliography{sample}

\end{document}